\documentclass[10pt, a4paper]{article}
\usepackage{lrec}
\usepackage{multibib}
\newcites{languageresource}{Language Resources}
\usepackage{graphicx}
\usepackage{subcaption}
\usepackage{tabularx}
\usepackage{soul}
\usepackage{arabtex}
\usepackage[T1]{fontenc}

\usepackage{epstopdf}
\usepackage[latin1]{inputenc}

\usepackage{url}
\usepackage{listings}
\usepackage{color}
\definecolor{gray}{rgb}{0.4,0.4,0.4}
\definecolor{darkblue}{rgb}{0.0,0.0,0.6}
\definecolor{cyan}{rgb}{0.0,0.6,0.6}

\newcommand{\hide}[1]{}

\title{\textbf{{\it System Demonstration}\\MADARi: A Web Interface for Joint Arabic \\ Morphological Annotation and Spelling Correction}}

\name{Ossama Obeid, Salam Khalifa, Nizar Habash, \\
{\fontsize{12pt}{12pt}{\textbf{Houda Bouamor,$^\dagger$ Wajdi Zaghouani,$^\star$ Kemal Oflazer$^\dagger$}}}
}

\address{Computational Approaches to Modeling Language Lab, New York University Abu Dhabi, UAE \\
         $^\dagger$ Carnegie Mellon University in Qatar, Qatar,\
  $^\star$ Hamad Bin Khalifa University,  Qatar\\
         \{oobeid, salamkhalifa, nizar.habash\}@nyu.edu, \\hbouamor@qatar.cmu.edu, wzaghouani@hbku.edu.qa, ko@andrew.cmu.edu}

\abstract{
We introduce MADARi, a joint morphological annotation and spelling correction interface for Arabic text. 
Our framework provides intuitive interfaces for annotating text and managing the annotation process.
We describe  the motivation, design and implementation of this interface; and we present details from a user study working with this system. 
\\ \newline  }

\begin{document}

\maketitleabstract
\setarab
\novocalize

\section{Introduction}
%Corpus-based research has been the major driving force in recent natural language processing (NLP) developments.
%Computational Linguists use annotated corpora among other things, to observe and propose linguistic hypotheses, to optimize them and to finally evaluate them (or the approaches based on those rules).
%Manual annotation is a tedious and a difficult task. Ideally, a good annotation tool would allow you to easily input an annotation schema for entities and relationships. 

Annotated corpora have been vital for research in the area of natural language processing (NLP).These resources provide the necessary training and evaluation data to build automatic annotation systems and benchmark them.
The task of human manual annotation, however, is rather difficult and tedious; and as such a number of annotation interface tools have been created to assist in such effort.
These tools tend to be specialized to optimize for specific tasks such as spelling correction, part-of-speech (POS) tagging, named-entity tagging, syntactic annotation, etc.
Certain languages bring additional challenges to the annotation task. Compared with English,  Arabic annotation introduces a need for diacritization of the diacritic-optional orthography, frequent clitic segmentation, and a richer POS tagset.
Although the goal of  language-independence is something most researchers and interface developers keep in mind, it is rather hard to  achieve without a tradeoff with utility and efficiency.

In this paper, we focus on a tool targeting Arabic dialect morphological annotation.  Arabic dialects introduce yet more complexity than standard Arabic in that the input text has noisy orthography. 
For example, the last word in the sentence used as example in Figure~\ref{fig:madari}.(a), <wyaabuwhA-Al_hliyj> {\it wyAbwhAAlxlyj}\footnote{All transliteration is in the Buckwalter scheme \cite{arabic-transliteration}.} involves two spelling errors (a word merge and character replacement) which can be corrected as <wjaabuwhA Al_hliyj> {\it wjAbwhA Alxlyj} `and they brought it to the Gulf'.  Furthermore, the first of the two corrected words includes two clitics that when segmented produce the form:   <hA>+ <jaabuwA> +<w> {\it w+ jAbwA +hA} `and+ they-brought +it'.

Previous work on Arabic morphology annotation interfaces focused either on the problem of manual annotations for POS tagging, or diacritization, or spelling conventionalization. 
In this paper we present a tool that allows one to do all of these tasks together, eliminating the possibility of error propagation from one annotation level to another.
Our tool is named MADARi\footnote{<madaary> {\it madAriy} means `my orbit' in Arabic.} after the project under which it was created:  Multi-Arabic Dialect Annotations and Resources (MADAR).
%cite MADAR paper. in camera ready.
% we can add links to tools for syntax annotation; particularly Palmyra.

Next, we present related work to this effort. In Section~3, we discuss the MADARi task description and design concerns.  In section~4 and~5, we discuss the annotation and management interfaces, respectively. Section~6 presents some details on a user study of working with MADARi.

\section{Related Work}
Several annotation tools and interfaces were proposed for many languages and to achieve various annotation tasks such as the general purpose annotation tools BRAT \cite{Stenetorp:2012}, WebAnno~\cite{Webanno}.
For task specific annotation tools, we can cite the post-editing and error correction tools such as the work of \newcite{aziz+2012:pet}, \newcite{Stymne:2011:BTE:2002440.2002450}, \newcite{conflrec}, and \newcite{dickinson}.
For Arabic, there are several existing annotation tools,
however, they are designed to handle a specific NLP task and it is not easy to adapt them to our project.
We can cite tools for semantic annotation such as the work of \newcite{saleh2009aratation} and \newcite{el2014proposed} and the work on dialect annotation by \newcite{benajiba2010web} and \newcite{Diab10colaba:arabic}.
\newcite{AttiaRA09} built a morphological annotation tool and more recently MADAD \cite{ALTWAIRESH16.619}, a general-purpose online collaborative annotation tool for Arabic text was designed during a readability assessments project.
In the COLABA initiative \cite{Diab10colaba:arabic}, the authors built tools and resources to process Arabic social media data such as blogs, discussion forums, and chats.
Above all, most, if not all of these tools are not designed to handle the peculiarities of the dialectal Arabic, which is a very specific task.
Moreover, the existing tools, do not provide facilities for managing thousands of documents and they often do not permit the distribution of tasks to tens of annotators while evaluating the inter-annotator agreement (IAA).
Our interface borrows ideas from three other annotation tools: DIWAN, QAWI, and MANDIAC.
Here we describe each of these tools and how they have influenced the design of our system.

\paragraph{DIWAN}
DIWAN is an annotation tool for Arabic dialectal texts \cite{AlShargi:2015}.
It provides annotators with a set of tools for reducing duplicate effort including the use of morphological analyzers to precompute analyses, and the ability to apply analyses to multiple occurrences simultaneously.
However it requires installation on a Windows machine and the user interface is not very friendly to newcomers.

\paragraph{QAWI}
The QALB Annotation Web Interface (QAWI) first introduced the concept of token-based text edits for annotating parallel corpora used in text correction tasks~\cite{Obeid:2013,zaghouani2014large}.
It allowed for the exact recording of all modifications performed by the annotator which previous tools did not.
As we show later on, we utilize this token-based editing system for minor text corrections that transform text of a given dialect into the appropriate CODA format.

\paragraph{MANDIAC}
MANDIAC \cite{Obeid:2016} utilized the token-based editor used in QAWI to perform text diacritization tasks.
More importantly, it introduced a flexible hybrid data storage system that allows for adding new features to the annotation front-end with little to no modifications to the back-end.
Our annotation system utilizes this design to provide the same utility.

\section{MADARi Design}

\paragraph{Task Description}
The MADARi interface will be used by human annotators to create a morphologically  annotated corpus of Arabic text. The text we work with comes from social media and is highly dialectal and as such, it has a lot of spelling errors.  The annotators will carefully correct the spelling of the words in the text and also annotate the words' morphology.  The in-context morphology annotation includes tokenization, POS tagging, lemmatization and English glossing.  

\paragraph{Desiderata}
In order to manage and process the annotation of the large scale dialectal Arabic corpus, we needed to create a tool to streamline the
annotation process.   

The desiderata for developing the MADARi annotation tool include the following:
 
\begin{enumerate}
  \setlength{\itemsep}{0pt}
  \setlength{\parskip}{0pt}
  \setlength{\parsep}{0pt}
\item  No installation time and very minimal requirements on the annotators.
\item  The tool must allow off-site data management of documents to allow annotation leaders to assign and grade documents from anywhere in the world and to allow hiring annotators anywhere in the world.
\item  The tool must allow easily customizable POS tag sets by annotation leads.
\item  The tool must allow easy access to other user annotations of similar texts.
\item  The tool must allow for easy navigation between spelling changes and morphological disambiguation. 
\end{enumerate}

\paragraph{Design and Architecture}
The design of our interface borrows heavily from the design of MANDIAC \cite{Obeid:2016}.
In particular, we utilized the client-server architecture, as well as the flexible hybrid SQL/JSON storage system used by MANDIAC.
This allows us to easily extend our annotation interface with minor changes, if any, to the back-end.
Like, DIWAN and MANDIAC, we also utilize MADAMIRA \cite{MADAMIRA:2014}, a state-of-the-art morphological analyzer for Arabic to precompute analyses.

%%%%%%%%%%%%%%%%%%%%%%%

\begin{figure*}
  \centering

    \begin{subfigure}{1\textwidth}
        \centering
        \includegraphics[width=0.75\textwidth]{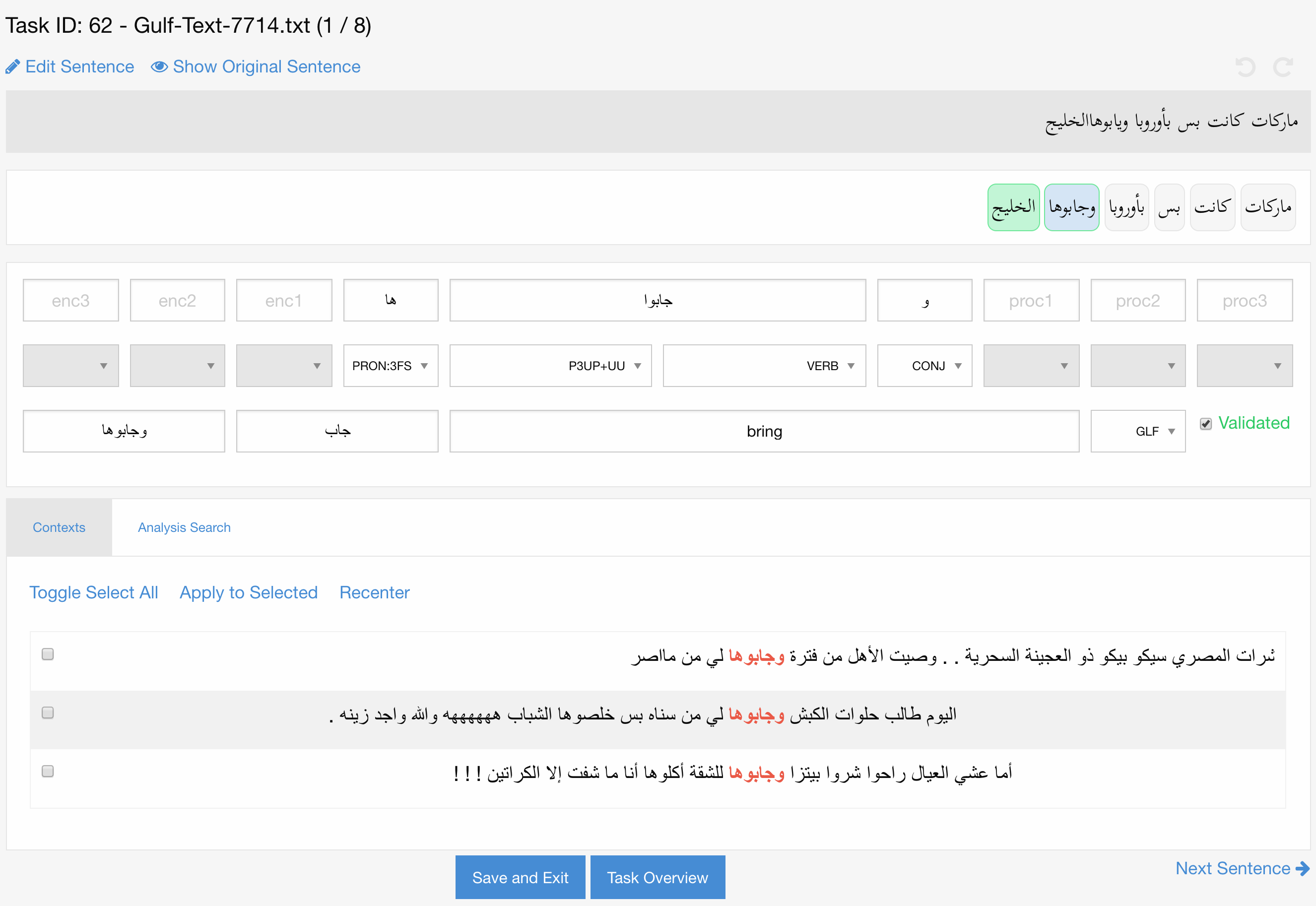}
        \caption{Full view of the MADARi annotation interface.}
        \label{madari:sfig1}
    \end{subfigure}
    \par\bigskip
    \par\bigskip

    \begin{subfigure}{1\textwidth}
        \centering
        \includegraphics[width=0.75\textwidth]{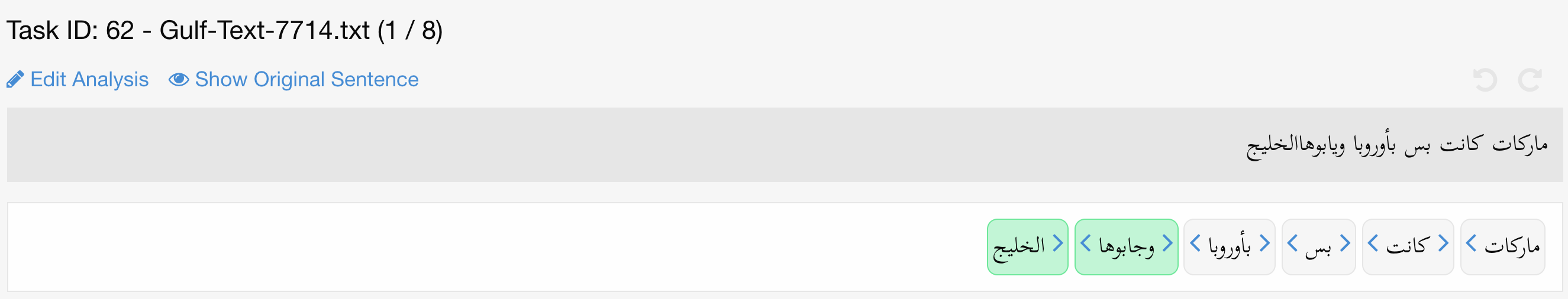}
        \caption{Text edit mode.}
        \label{madari:sfig2}
    \end{subfigure}
    \par\bigskip\
    \par\bigskip

    \begin{subfigure}{1\textwidth}
        \centering
        \includegraphics[width=0.75\textwidth]{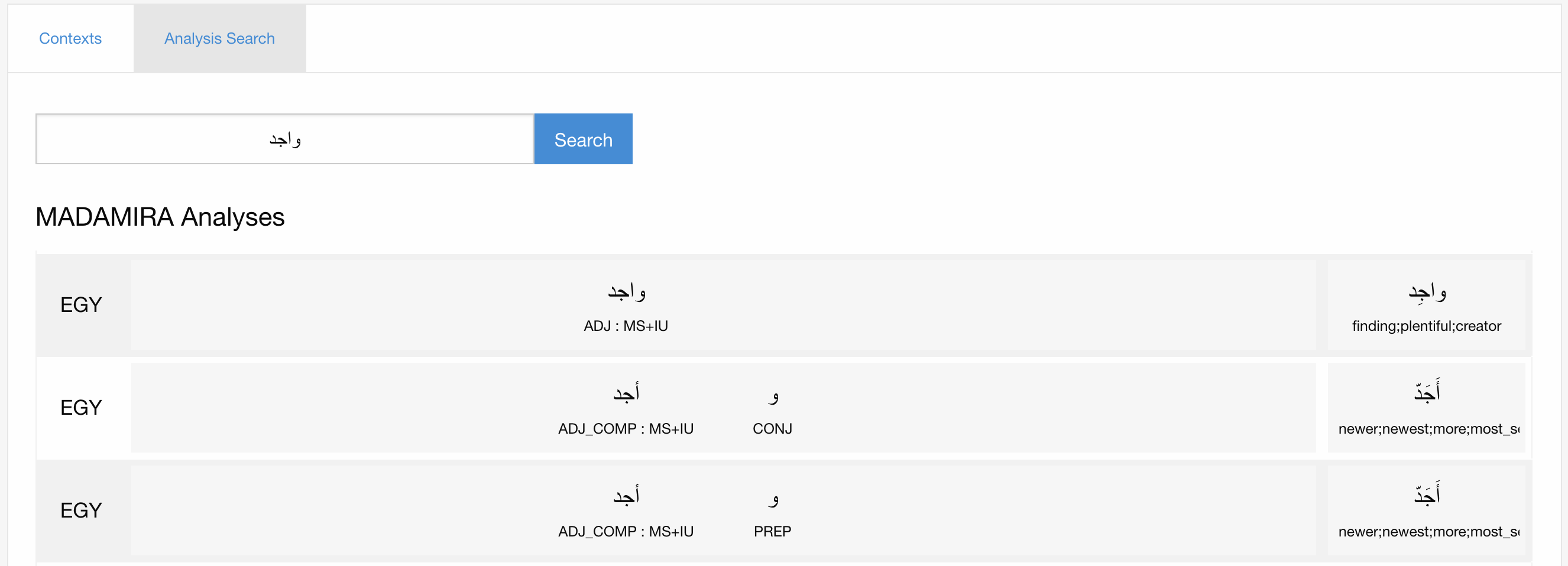}
        \caption{Analysis search panel.}
        \label{madari:sfig3}
    \end{subfigure}
    \par\bigskip
    \par\bigskip
    \caption{The MADARi Annotation Interface}
    \label{fig:madari}

\end{figure*}

%%%%%%%%%%%%%%%%%%%%%%%

\section{Annotation Interface}
The Annotation Interface (Figure \ref{madari:sfig1}) is where annotators perform the annotation tasks assigned to them.
Here we describe the different components and utilities this interface provides.

\paragraph{Text Editing}
Annotators are able to edit a sentence anytime during the annotation process.
This is primarily used to make sure all text is in the CODA format of the dialect of choice.
We adopted the same token-based editing system used by QAWI.
Our token-based editor (Figure \ref{madari:sfig2}) only allows for modifying, splitting, and merging tokens where QAWI also allows for adding and deleting tokens as well as moving tokens around.
The operations we allow are sufficient for CODA formatting without allowing the text to be changed substantially.

\paragraph{POS Tagging}
The essential component of our interface is the POS tagging system.
Here, all words are annotated in their tokenized form which divides a word into its base word, enclitics, and proclitics.
Each of these are assigned a POS tag as well as a morphological feature where applicable.
Annotators also assign the gloss and lemma for each word.
For the convenience of annotators, we provide precomputed values for each field using MADAMIRA's morphological analyzers.

\paragraph{Utilities}
We have added utility features to make the annotation process easier and more efficient for annotators.
Basic utilities include undo and redo buttons, access to the original text for reference, and color-coding edited tokens for quick navigation as seen in Figure \ref{madari:sfig1}.
We also allow annotators to update multiple tokens with the same orthography instantaneously.
Additionally, we provide annotators with a search utility to look up previously submitted annotations of the same word as well as query MADAMIRA for out-of-context analyses in different dialects in real-time (Figure \ref{madari:sfig3}).

\section{Management Interface}
The Annotation Management Interface enables the lead annotator to easily manage and organize the whole annotation process remotely and concurrently. The management interface contains: (a) a user management tool for creating
new annotator accounts and viewing annotator
progress; (b) a document management tool for
uploading new documents, assigning them for annotation,
and viewing submitted annotations; and
(c) a monitoring tool for viewing overall annotation
progress; (d) an inter-annotator agreement (IAA) evaluation tool to compare the annotations produced by each annotator to a gold reference in order to monitor the quality of the annotations; and (e) a data repository and annotation export feature.

\section{User Study}
Our tool is being used as part of an ongoing annotation project on Gulf Arabic (forthcoming).
In this paper, we describe the experience of one annotator who has done annotations in different settings previously.
The annotator morphologically disambiguated 80 sentences totaling in 1,355 raw tokens of Gulf Arabic text.

We noted that the annotator preferred, based on her experience, to convert the orthography of the text to CODA first, which made the disambiguation task more efficient.

It took about 52 minutes to complete this task (corresponding to a rate of 1,563 words/hour). The annotator made a few minor fixes later on, which is an advantage of our tool to minimize error propagation.
The total number of words that were changed from the raw tokens to CODA was 288 (21\%).
Changes were mostly spelling adjustments and the rest is word splitting (44 cases or 15\% of all changes) and no merges. The final word count is 1,398 words. 

Following the CODA conversion, the annotator worked on tokenization, POS tagging, lemmatization and English glossing.
This more complex task took around 6 hours (at a rate of 277 words/hour).
This makes the cumulative time spent to finish the spelling adjustment and the full disambiguation tasks for this set of data about 7 hours (at a rate of 200 words/hour).

Since the tool provides initial guesses for all the annotation components, the annotator was able to use many of the valid decisions as is, and modify them in other cases.
In the event of a word split, the tool currently removes the raw word predictions, but the analysis search utility allows fast access to alternatives to select from.
% TODO: in camera ready report combined changes with alignments -- TBD
% 
We compared the final tokenization, POS tag and lemma choices to the ones suggested by the tool on the CODA version of the text. We found that the tool gave correct suggestions 74\% of the time on tokenization, 69\% of the time on baseword POS tags and 70\% of the time on lemmas.

The annotator indicated that their favorite utilities were the ability to annotate multiple tokens of the same type in different contexts simultaneously,
and the ability to use the `Analysis Search' box to annotate multiple fields simultaneously.

%\footnote{these numbers are calculated posthoc using the correct CODA forms as the starting point. 
%We opted for this calculation to 
% TODO: Ossama - allow retagging.

\section{Conclusion and Outlook}
We presented an overview of our web-based
annotation framework for joint morphological annotation and spelling correction of Arabic. We plan to release the tool and make it freely available to the research community so it can be used in other related annotation tasks. In the future, we will continue extending the tool to work on different dialects and genres of Arabic.  

\section*{Acknowledgments}
This publication was made possible by grant NPRP7-290-1-047  from the Qatar National Research Fund (a member of the Qatar Foundation).  The statements made herein are solely the responsibility of the authors.

\section*{Bibliographical References}
\label{main:ref}

\bibliographystyle{lrec}
\bibliography{ALLBIB-2.5}

%%%%%%%%%%%%%%%%%%%%%%%

% \begin{figure*}
%   \centering
% \begin{tabular}{cc}

% (A) &    \includegraphics[width=0.8\textwidth]{madari_full.png}\\\\\hline\\

% (B) &    \includegraphics[width=0.8\textwidth]{madari_textedit.png}\\\\\hline\\

% (C) &   \includegraphics[width=0.8\textwidth]{madari_analysis.png}\\\\\hline\\
% \end{tabular}
% \caption{(A) Full view of the MADARi annotation interface. (B) Text edit mode. (C) Analysis search panel.}
% \label{fig:annot_interface}
% \end{figure*}

%%%%%%%%%%%%%%%%%%%%%%%%%%%

% \section{Language Resource References}
% \label{lr:ref}
% \bibliographystylelanguageresource{lrec}
% \bibliographylanguageresource{xample}

\end{document}